\documentclass{article} 
\usepackage{iclr2020_conference,times}
\usepackage[english]{babel}
\usepackage[utf8]{inputenc}

\usepackage{amsmath,amsfonts,bm}

\def\eqref#1{equation~\ref{#1}}

\def\1{\bm{1}}

\DeclareMathAlphabet{\mathsfit}{\encodingdefault}{\sfdefault}{m}{sl}
\SetMathAlphabet{\mathsfit}{bold}{\encodingdefault}{\sfdefault}{bx}{n}

\usepackage{hyperref}
\usepackage{url}

\usepackage{booktabs} 
\usepackage{graphicx}

\usepackage{pifont}
\newcommand{\xmark}{\text{\ding{55}}}

\usepackage{multirow,bigdelim}

\def\remark#1{}
\def\lukas#1{}
\def\david#1{}
\def\timo#1{}
\def\todo#1{}
\def\question#1{}

\newcommand{\arne}{ARNe}

\title{Attention on Abstract Visual Reasoning}
\iclrfinalcopy

\author{
Lukas Hahne
 \qquad Timo Lüddecke \qquad Florentin Wörgötter \qquad David Kappel\\[3mm]
Correspond to: \texttt{l.hahne@stud.uni-goettingen.de}\\[3mm]
Bernstein Center for Computational Neuroscience \\
III Physikalisches Institut-Biophysik, Georg-August Universität\\
Göttingen, Germany
}

\begin{document}

\maketitle

\begin{abstract}
Attention mechanisms have been boosting the performance of deep learning models on a wide range of applications, ranging from speech understanding to program induction.  However, despite experiments from psychology which suggest that attention plays an essential role in visual reasoning, the full potential of attention mechanisms has so far not been explored to solve abstract cognitive tasks on image data. In this work, we propose a hybrid network architecture, grounded on self-attention and relational reasoning. We call this new model \emph{Attention Relation Network} (ARNe). ARNe combines features from the recently introduced Transformer and the Wild Relation Network (WReN). We test ARNe on the Procedurally Generated Matrices (PGMs) datasets for abstract visual reasoning. ARNe excels the WReN model on this task by 11.28~ppt. Relational concepts between objects are efficiently learned demanding only $35\%$ of the training samples to surpass reported accuracy of the base line model. Our proposed hybrid model, represents an alternative on learning abstract relations using self-attention and demonstrates that the Transformer network is also well suited for abstract visual reasoning.
\end{abstract}

\section{Introduction}

Psychological models of human intelligence identify different manifestations of intellect. \textsc{Fluid Intelligence} describes the ability to adapt to new problems and situations without relating to previous learning outcomes \citep{hagemann2016differentielle}. This ability is considered as one of the most important aspects for learning and is essential for solving higher cognitive tasks \citep{Jaeggi}. In order to excel in this type of intelligence, cognitive capabilies such as figural relations, memory span and inductive thinking are decisive. Fluid intelligence also paves the ground for \textsc{abstract reasoning}, the ability to use symbols instead of concrete objects. Furthermore, empirical data show that attention and fluid intelligence are strongly interlinked \citep{stankov1983attention,schweizer2010relationship,ren2013sources}. Subjects that perform bad on attention tasks are also more likely to show deficits in abstract reasoning and fluid intelligence \citep{ren2012does,ren2013sources}.

\begin{figure}
    \centering
    \includegraphics[width=\textwidth]{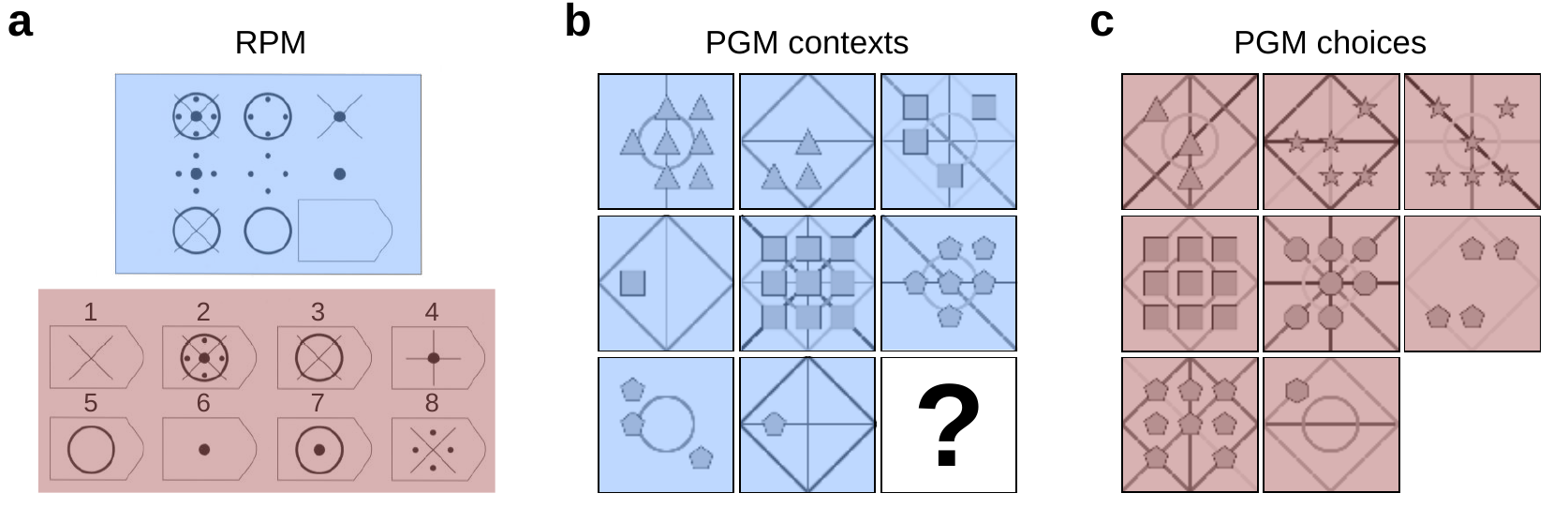}
    \caption{\textbf{Datasets for visual reasoning.} \textbf{a:} Sample from the RPM dataset \citep{bilker2012development}. Context: blue, choices: red. Correct choice: 1. Implicit rule: Subtraction along rows and columns. \textbf{b, c:} Sample from the PGM dataset \citep{barrett2018measuring}, contexts (\textbf{b}) and corresponding choices (\textbf{c}). The \textbf{?} in (\textbf{b}) should be replaced by a correct choice from (\textbf{c}). Correct choice: The hexagon in the last tile. Implicit rule: progression of shape types (number of edges) in each column. Colors added to enhance visualization. \remark{made a b c bold}}
    \label{fig:RPM-PGM}
\end{figure}

\textsc{Raven's Progressive Matrices} (RPM) is an established test method for intelligence, especially fluid intelligence and abstract reasoning \citep{bilker2012development}. It is a set of non-verbal tests showing several geometric objects arranged to a certain implicit rule (Figure~\ref{fig:RPM-PGM}a shows an example). The test subject has to complete the $3 \times3$ matrix by picking an object that matches the implicit rule. The \textsc{Procedurally Generated Matrices} (PGMs) \citep{barrett2018measuring} dataset is motivated by RPMs and synthetically generated for training neural networks. Its core feature comprises several relations between objects and one of their attributes, e.g. the number of an object of a certain type in each matrix panel. A single matrix can contain up to four such relations simultaneously. A PGM's context and possible answers are shown in Figures~\ref{fig:RPM-PGM}a~and~\ref{fig:RPM-PGM}b, respectively. Importantly, the PGM matrices are presented to a neural network as images such that the task requires a certain level of understanding of geometrical relations. The Wild Relation Network (WReN) uses different combinations of matrix elements as a mechanism for relational reasoning and is currently among the best performing solutions for the PGM dataset. However, despite the close relationship between attention and reasoning, the WReN is lacking a mechanism for attention.

Apart from the visual domain, it was recently shown that attention is also essential for another cognitive task that fundamentally relies on abstract reasoning: \emph{language understanding}. The Transformer, an attention-based neural network, was introduced to improve machine translation and transduction \citep{vaswani2017attention}. The Transformer embodies a self-attention mechanism to relate parts of its input with each other for writing output. Since this form of attention and abstract reasoning is vital for fluid intelligence we hypothesize that a Transformer augmented with relational reasoning capabilities can perform well on the PGM task.

Another special trait of human intelligence is the ability to infer rules with little supervision. Often, only few samples suffice for people to grasp the idea of tasks such as visual reasoning. In contrast to that, current deep network architectures are trained on huge quantities of data, e.g. the PGM dataset contains more than 1.2 million labelled samples. To see if self-attention also enhances this capability of deep networks we investigate the sample efficacy by training our proposed model only on a fraction of the total available data samples.

Our contribution to the field of machine learning is three-fold:
\begin{itemize}
    \item We introduce the Attention Relation Network (ARNe) that combines features from the WReN and the Transformer network and can be directly trained on visual reasoning tasks.
    \item We evaluate ARNe on the PGM task and show that it significantly outperforms the current state-of-the art \citep{steenbrugge18} by 11.28~ppt.
    \item We demonstrate that ARNe is very sample-efficient and achieves its peak performance with only 35\% of the full PGM dataset.
\end{itemize}

This paper is organized as follows. In Section~\ref{sec:related-work} we discuss related approaches and training datasets. In Section~\ref{sec:arne} we introduce the ARNe architecture and in Section~\ref{sec:experiments} we show our simulation results. We conclude in Section~\ref{sec:conclusion}.

\section{Related Work}
\label{sec:related-work}

\paragraph{Language Modelling}
Language modelling, i.e. predicting the likelihood of future words given a set of contextual words, is a core task in natural language processing (NLP). 
While neural approaches to language modelling have some tradition \citep{bengio03}, the success of the word2vec \citep{word2vec} has revived and boosted the interest in neural networks for this task.
While word2vec defines fixed vectors for words, recently, several approaches were proposed that condition word representations on their context \citep{howard18, devlin18, peters18}.
In particular attention-based models have emerged as the standard tool for language modelling as well as several downstream tasks, such as question answering \citep{squad} and translation \citep{vaswani2017attention}.
Starting with the seminal Transformer \citep{vaswani2017attention}, follow-up models were successively improved \citep{devlin18} and trained on larger text datasets achieving remarkable success across various NLP tasks \citep{radford2019better}.
Besides the work in language modelling, in concurrent work, Transformer-based models have been employed for visual question answering \citep{li2019visualbert}.
In this work we take inspiration from the model design of the attention Transformer. However, instead of applying it in a language modelling task, we use it for abstract reasoning.

\paragraph{Abstract Reasoning}
Recently, there has been growing interest in abstract reasoning indicated by the introduction of datasets operating on textual data, e.g. babl \citep{weston2015towards}, as well as on visual data, involving CLEVR \citep{johnson2017clevr}, FigureQA \citep{figureqa} and NLVR \citep{suhr2017corpus}.

While classic recurrent neural networks have often been used for these tasks, it has been found that novel architectures such as attention and memory are better suited to address abstract reasoning. Attention mechanisms are a method to simplify complex information by attending to different locations of the given data. This is an advantageous procedure to let a neural network determine a specific reasoning operation on these selected locations. Memory is needed to preserve information in tasks where multiple consecutive reasoning operations are required. In LSTM \citep{hochreiter1997long} for example, the hidden state is protected by gates from being overwritten by new input.
Models incorporating attention and memory involve the Neural Turing Machine \citep{graves14} which evolved into the Differentiable Neural Computer \citep{graves2016hybrid}. 
Dynamic Memory Networks \citep{xiong2016dynamic} address both, visual and textual question answering.
The recently introduced Memory, Attention and Composition (MAC) Network \cite{hudson2018compositional} achieved excellent scores in the CLEVR benchmark by using multiple explicit reasoning steps. Recently, a reasoning module relying on attention \cite{zambaldi2018deep} was designed to be employed in a reinforcement learning framework.

Our work differs by measuring fluid intelligence by understanding visual patterns rather than solving referential expressions (to which part of an image belongs a certain word) which is required in CLEVR, FigureQA, NLVR and many visual question answering tasks. The former task is considered in \citep{barrett2018measuring} and \citep{steenbrugge18} with which we compare our proposed method.


\def\y{\mathbf{y}}
\def\p{\mathbf{p}}
\def\d{\mathbf{d}}
\def\a{\mathbf{a}}

\def\oo{\mathbf{o}}
\def\pp{\mathbf{\hat{p}}}
\def\ap{\mathbf{\hat{a}}}
\def\x{\mathbf{x}}
\def\c{\mathbf{c}}
\def\d{\mathbf{d}}
\def\chiB{\boldsymbol{\chi}}

\section{Attention Relation Network (ARNe)}
\label{sec:arne}

\begin{figure}
    \centering
    \includegraphics[width=\textwidth]{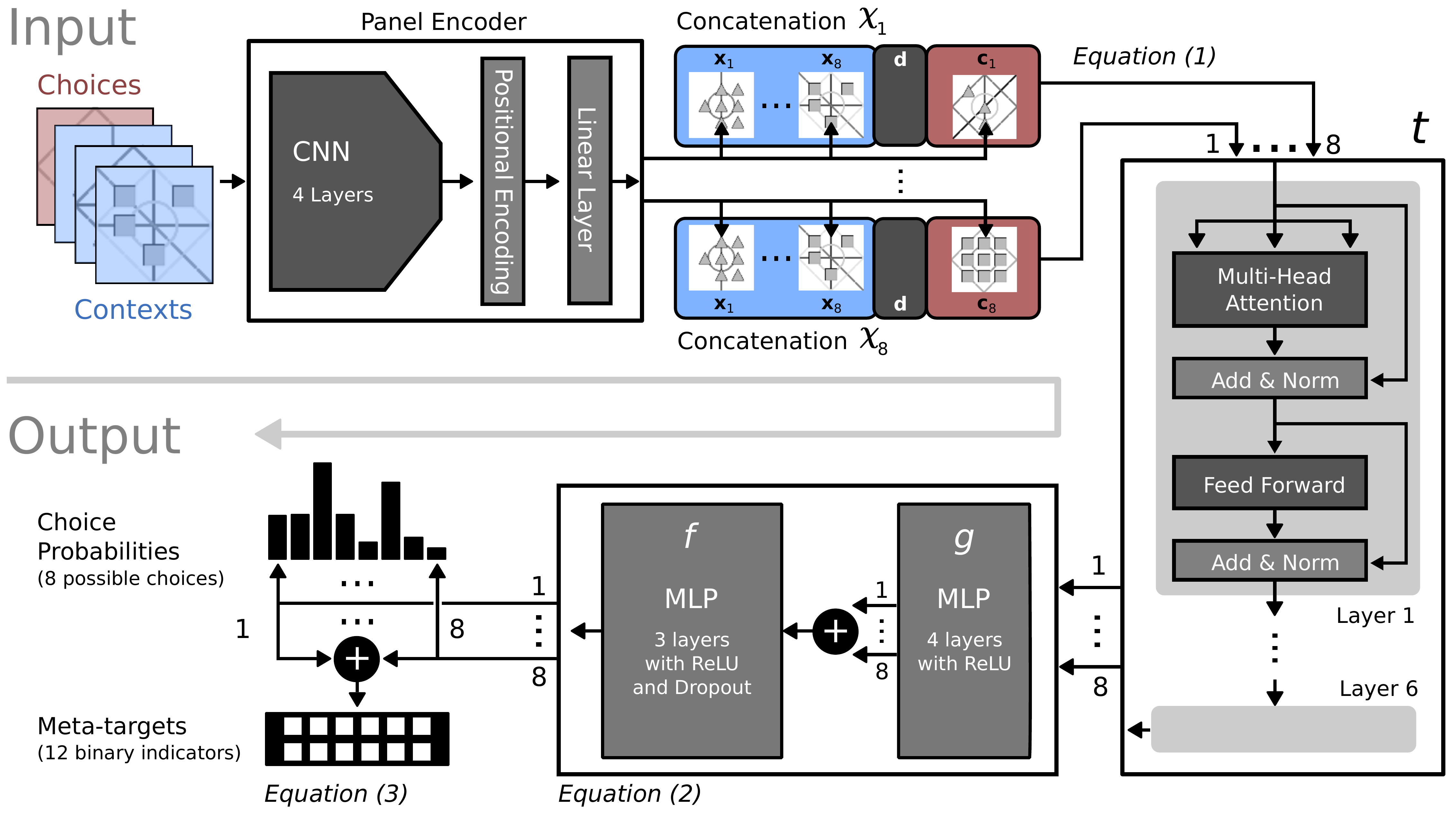}
    \caption{Overview of the \arne{} network for abstract visual reasoning.}
    \label{fig:arne}
\end{figure}

In this section we describe the Attention Relation Network (\arne), which combines techniques from language modelling to abstract reasoning.
It takes eight contextual panels and eight choice panels from the PGM, of which one fits to abstract relational rule implicitly determined by the context panels. The model output are logits of a corresponding fitness probability for each choice panel and logits for corresponding probabilities of the rules embedded in a PGM. The model is illustrated in Figure~\ref{fig:arne}.


First, representations for context panels, denoted by $\x_i$, and representations for choice panels, denoted by $\c_k$, are generated using a shared \emph{Convolutional Neural Network} (CNN). $\x_i$ denotes the $i$-th context panel feature, $\c_k$ denotes the $k$-th choice panel feature. This network has the same hyperparameters as the CNN in the Wild Relation Network (WReN). A one-hot positional encoding indicating one out of nine possible positions within the panel grid is concatenated to each extracted panel feature and the resulting vector is projected to obtain a final representation for each panel (see \emph{panel encoder} and \emph{concatenation} blocks in Figure~\ref{fig:arne}).

Sequences $\chiB_k$ of length $N$ are composed of the context and choice panel representations. Optionally, a learnable deliminiter $\d$ which has the same number of dimensions as the panel representations can be included between contexts and choice. Thus, sequences of $N=9$ or $N=10$ elements are obtained:
\begin{align}
\chiB_k = \left( \x_1, \x_2, \ldots, \x_8, \d, \c_k \right) \label{eq:arne_chi}\;.
\end{align}
We generate a sequence $\chiB_k$ for all eight choices. A multi-step attention network inspired by the encoder of the Transformer model $t$ \citep{vaswani2017attention}, processes this sequence using the self-attention mechanism for abstract reasoning. All activations are accumulated before the network's output is generated by the MLP $f$.
\begin{align}
 \oo_k = f \left(\sum_{i=1}^{N} g \left( t \left(\chiB_k \right) \right)_i \right)\;, \label{eq:ARNe}
\end{align}
where $g$ denotes the output of the 4-layer MLP (see $f$ and $g$ decoder blocks in Figure~\ref{fig:arne}).
The vector $\pp \in \mathbb{R}^8$ denotes the logits of the corresponding probability distribution over the choices and the matrix $\mathbf{\hat{A}} \in \mathbb{R}^{8 \times 12}$ indicates the logits of the corresponding presence probabilities of rules imposed by the panels where $\ap_k \in \mathbb{R}^{12}$ is defined as a transposed row vector of this matrix. The construction of these meta-targets is explained subsequently.

\begin{table}
    \centering
    \caption{Modules and their corresponding parameters of \arne{}. \label{tab:ARNe}}
    \vspace{3mm}
    \begin{tabular}{lrlll} 
    \toprule
     \multicolumn{3}{l}{\textbf{Module}}&& \textbf{Parameters}\\ 
     \midrule
     \textbf{Panel Encoder} \\[2mm]
     CNN &\ldelim\{{3}{20pt}[] & 2D-Convolution & \rdelim\}{3}{20pt}[x4] & $3 \times 3 \times 32$, $\text{padding}=1, \text{stride}=2$ \\ 
      & &Batchnorm &  & \\ 
      & &ReLU & &\\[1mm]
      Linear layer && Embedding && $\frac{\text{H} \cdot \text{W}}{256} \cdot 32 + 9 \times 512 $ \\
      \midrule
     \textbf{Transformer} \\[2mm]
      $t$  && Transformer encoder  &  x6& see Table~\ref{tab:model_encoder_transformer}\\[1mm]
     $g$ &\ldelim\{{2}{20pt}[]& linear layer &  \rdelim\}{2}{20pt}[x4] & each $512 \times 512$\\
     && ReLU & & \\[1mm]
     $f$ &\ldelim\{{3}{20pt}[]&  (Dropout) & \rdelim\}{3}{20pt}[x3] &   p = $0.5$ (only for second layer)\\
     && linear layer && $512 \times 256$, $256 \times 256$ and $256 \times 13$ \\
     && ReLU && \\ 
    \bottomrule
    \end{tabular}
\end{table}
The network returns a matrix $\oo \in \mathbb{R}^{8 \times 13}$, which contains both, logits for the prediction of a choice panel and logits for the prediction of underlying patterns across the PGM\\[2mm]
\begin{equation}
        \begin{pmatrix}
        \hat{p}_1 & \ap_1^T\\
        \vdots & \vdots \\
        \hat{p}_8  & \ap_8^T
        \end{pmatrix}   =      
        \begin{pmatrix}
        \oo_1 \\
        \vdots \\
        \oo_8
        \end{pmatrix} .
\end{equation}

\paragraph{Meta-targets}

\def\y{\mathbf{y}}
\def\p{\mathbf{p}}
\def\a{\mathbf{a}}
\def\pp{\mathbf{\hat{p}}}
\def\ap{\mathbf{\hat{a}}}
\def\x{\mathbf{x}}
\def\c{\mathbf{c}}
\def\d{\mathbf{d}}
\def\chiB{\boldsymbol{\chi}}

In the PGM dataset, the layout and appearance of the nine panels follow an implicit rules. Each rule is represented by a triplet containing an object type, object attribute and relation type. An example rule could be (shape, size, progression) to describe a pattern of triangles with increasing size.
Up to four different relational rules (i.e. four triplets) may occur simultaneously in a single PGM. Meta-targets are created by aggregating all binary encoded rules with an OR operation resulting in a 12 dimensional meta-target vector.

In order to identify correct panels, a successful model should be able to infer the underlying construction principle reliably. To enhance rule prediction in the model, we incorporate the auxiliary information encoded in the meta-targets in the loss function.
Since each choice panel results in a prediction $\ap_k$ of a meta-target, the predictions need to be aggregated by a sum over choice panels, implying an OR relation between the elements \citep{barrett2018measuring}.

\paragraph{Loss}

The loss $\mathcal{L}$ is defined by a weighted combination of finding the right choice $\p$ and detection of the correct logical pattern $\a$ \citep{barrett2018measuring}:
\[
\mathcal{L} = \text{CE}(\p, \pp) + \beta \cdot \text{BCE}(\a, \sum_{k=1}^8 \ap_k)\;,
\]
with CE and BCE being cross entropy and binary cross entropy.
The parameter $\beta$ controls the influence of the meta-targets, i.e. for $\beta=0$ the meta-targets are ignored.
By enforcing correctness of the meta-targets we enable richer gradients that identify better parameters.


\paragraph{Implementation}

The learning is carried out using an adam \cite{kingma2014adam} optimizer with a batch size of $64$ and an initial learning rate of  $0.5 \cdot 10^{-4}$ using a learning rate scheduler with exponential decay. We apply early stopping \citep{Goodfellow-et-al-2016} with a patience of three epochs.
%
%
%
%
%
%
%
Parameters of the individual components of our method are presented Table~\ref{tab:ARNe}.
For further details we refer to the appendix or the implementation of our approach which is available here:\\
\texttt{http://hidden for blind review. Will be disclosed upon acceptance}

\section{Experiments}
\label{sec:experiments}
    
    \begin{table}[]
        \caption{PGM accuracy by previous methods and our model (ARNe). Accuracy of WReN as reported in \citep{NIPS2017_7082} and our implementation. MAC model included for comparison.}
        \vspace{3mm}
        \centering
        \small
        \begin{tabular}{llr}
             \toprule
             \textbf{Model} & & \textbf{Accuracy [\%]} \\
             \midrule
             MAC (our implementation) & &  12.6 \\
             VAE-WReN \citep{steenbrugge18} & $\beta=4$ & 64.2 \\
             WReN \citep{NIPS2017_7082} & $\beta=10$ & 76.9   \\
             WReN (our implementation) & $\beta=10$ & 79.0   \\
             WReN-MAC & & 79.6   \\
             \textbf{ARNe (our implementation)} & $\beta=10$ & \textbf{88.2} \\
             \bottomrule
        \end{tabular}
        \label{tab:sota}
    \end{table}

We evaluate ARNe on the PGM dataset \citep{pgm_github_deepmind}. Each PGM input sample consists of $160 \times 160$ pixel images. In total, there are 16 panels in every sample, eight of which define the context and the remaining eight define possible choices (see Figure~\ref{fig:RPM-PGM}). Similar to RPMs, a correct panel has to be chosen that matches the implicit relational rules encoded in the context panels. PGM includes meta-targets in the form of 12-bit feature vectors that denote relation, object and attribute types. The rules that underlie each sample are composed of 1 to 4 relational rules, chosen from the set (\emph{Progression},  \emph{AND}, \emph{OR}, \emph{XOR}, \emph{Consistent Union}). Figure~\ref{fig:RPM-PGM}b,c shows an example of the \emph{Progression} rule.

We trained ARNe on the $1.2 \times 10^6$ training samples from the PGM dataset where early stopping terminated training after 45 epochs. After training, ARNe detected answer panels with an accuracy of $88.18\%$ and auxiliary data with $98.72\%$. F1-score reached $0.9801$. For comparison, we included in Table~\ref{tab:sota} results for WReN \citep{NIPS2017_7082} and VAE-WReN \citep{hudson2018compositional} as baseline. These models were significantly outperformed by ARNe. It is noteworthy that our re-implementation of WReN also outperformed \citep{NIPS2017_7082} by 2.1~ppt despite careful code validation (see Table~\ref{tab:sota} and Appendix~\ref{sec:reimplementation_wren}).

\begin{figure}[hb!]
    \centering
    \includegraphics[width=\textwidth]{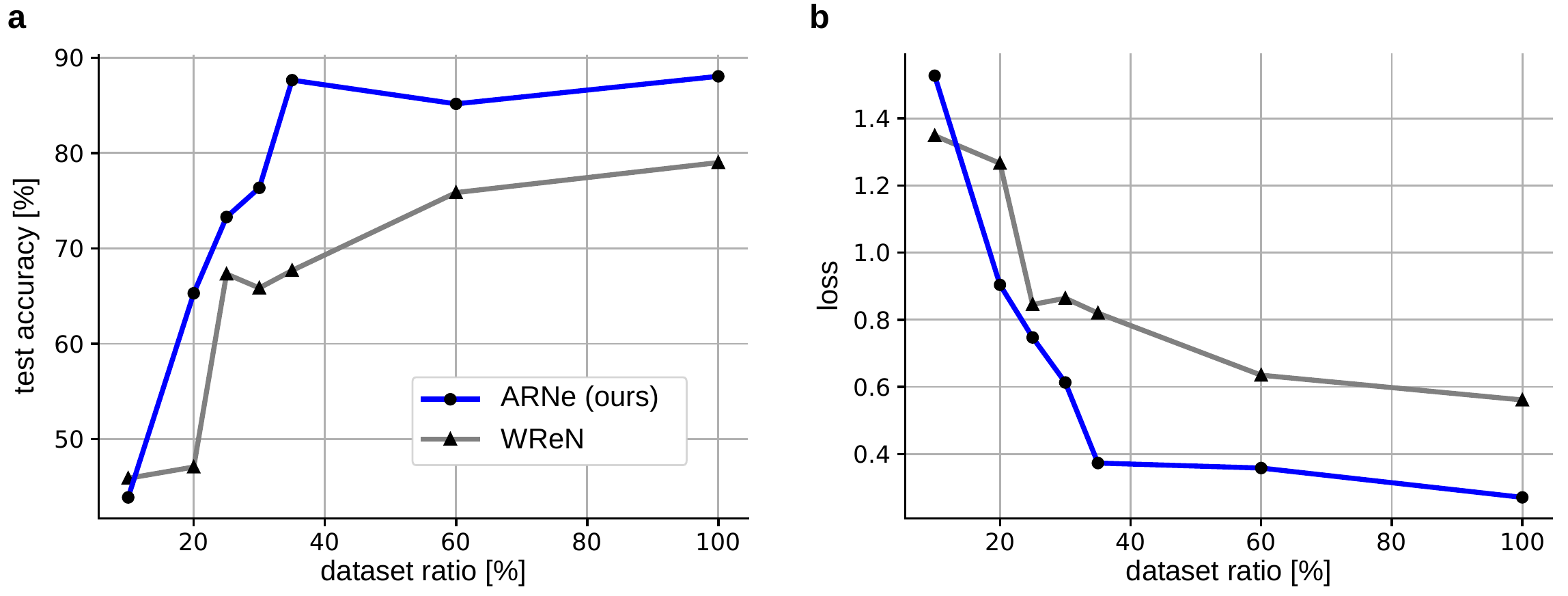}
    \caption{\textbf{Sample efficiency:} Test accuracy (\textbf{a}) and loss (\textbf{b}) after training on different numbers of samples (percent of original dataset size). Our implementation of WReN was used here (see Appendix~\ref{sec:reimplementation_wren}).}
    \label{fig:sample_efficiency}
\end{figure}

In addition, we included a comparison with the Attention and Composition (MAC) Network \citep{hudson2018compositional}. To the best of our knowledge this is the first time MAC is tested on the PGM dataset. To our surprise MAC did not perform significantly above chance level on this task. We also tested a version of MAC that uses a WReN at the input stage, similar to \arne{}. We call this augmented variant WReN-MAC. Our first results with this model suggest that also WReN-MAC does not reach the performance of ARNe. We ran tests on a reduced PGM dataset that used only 20\% of the full training set. WReN-MAC achieved 46.9\% test accuracy, about 10\% below that of \arne{} on the same dataset size (see Figure~\ref{fig:sample_efficiency}). We are running simulations with WReN-MAC on the full dataset but by the time of the submission of this paper these experiments were not concluded. Additional details on the implementation of WReN-MAC are provided in Appendinx~\ref{sec:wren-mac}.

\begin{table}
\caption{Meta-target prediction performance. \label{tab:binary_score_1}}
    \vspace{3mm}
\centering
\scriptsize
\begin{tabular}{rrrrr} 
\toprule 
 \textbf{Meta-target} & \textbf{Accuracy [\%]} & \textbf{Precision [\%]} & \textbf{Recall [\%]} & \textbf{F1-Score [\%]}\\ 
\midrule
Progression & $96.53$ & $94.07$ & $90.52$ & $92.26$\\
AND & $97.76$ & $96.92$ & $94.01$ & $95.44$\\
OR & $98.15$ & $98.53$ & $93.99$ & $96.21$\\
XOR & $97.59$ & $99.07$ & $91.23$ & $94.99$\\
Consistent Union & $96.87$ & $97.63$ & $91.79$ & $94.62$\\
\midrule
Shape & $99.82$ & $99.91$ & $99.81$ & $99.86$\\
Line & $99.98$ & $99.98$ & $99.98$ & $99.98$\\
\midrule
Size & $99.93$ & $99.89$ & $99.77$ & $99.83$\\
Type & $99.95$ & $99.91$ & $99.98$ & $99.94$\\
Position & $99.98$ & $99.91$ & $99.99$ & $99.95$\\
Number & $99.97$ & $99.92$ & $99.84$ & $99.88$\\
Color & $98.13$ & $99.01$ & $96.80$ & $97.89$\\
\bottomrule

\end{tabular}

\end{table}

To gain additional insights into the learning behavior of ARNe we clustered the test dataset by the number of relational dependencies. ARNe showed good performance on all types, performing best on samples with four (accuracy: $90.54\%$) and worst for samples with three (accuracy: $82.74\%$) relational rules. Next, we evaluated the ability of ARNe to predict the meta-targets. Table~\ref{tab:binary_score_1} shows individual performances for each of the 12 meta-targets. ARNe achieved high accuracy in all categories. Detection rates revealed an unconditional accuracy rate of above $90\%$ consistently across all meta-target types. 

Furthermore, we evaluate the generalization abilities of ARNe by the provided samples in the PGM dataset. Consistently with the scores on the neutral split, we find our method to outperform the wild relation network in Tab.~\ref{tab:generalisation_performance}.

\subsection{Sample Efficiency}

\newcommand{\SEbaseline}{$76.9\%$}
\newcommand{\SEbaselineReimplemented}{$79.00\%$}

\newcommand{\SEsplitDataset}{$20\%$}
\newcommand{\SEwrentransformerAccuracy}{$88.04\%$}
\newcommand{\SEWRENAccuracyDiff}{$9.04$~ppt}
\newcommand{\SEwrentransformerAccuracyDiff}{$0.40$~ppt}
    \newcommand{\SEsplitTransformerSizeSplit}{$35\%$}
\newcommand{\SEsplitTransformerBetterThanBaseline}{$30\%$}
\newcommand{\SEsplitTransformerBetterThanBaselineAccuracy}{$80.57\%$}

\newcommand{\SEwrentransformerLoss}{$??\%$}
\newcommand{\SEwrentransformerLossDiff}{$??\%$}

In order to evaluate the sample efficiency of ARNe, we trained the model subsequently with growing fractions of the training dataset. The validation and test sets were kept unchanged during the entire experiment. Prior to splitting of the training set, the dataset was shuffled and the random seed was kept fixed throughout all experiments. The results are presented in Figure~\ref{fig:sample_efficiency}. ARNe achieved higher test accuracy across all dataset ratios $\geq \SEsplitDataset{}$. Using a split size of \SEsplitTransformerSizeSplit{}, the accuracy reads $87.64\%$, already close to the value when trained on the full training set and $8.65$~ppt better than the WReN model at $100\%$ of the training dataset. The progression of the loss shows a similar behavior (Figure~\ref{fig:sample_efficiency}b).




\begin{table}
    \centering
    \caption{Generalisation results. No training results for WReN were available \citep{barrett2018measuring}. Values in the training, validation and test columns characterise the model's accuracy.\label{tab:generalisation_performance}}
    \small
    \begin{tabular}{lrrrr}
     \toprule
     Model & $\beta$ & Training $\left[\%\right]$ & Validation $\left[\%\right]$ & Test $\left[\%\right]$ \\
     \midrule
     WReN \citep{barrett2018measuring} & 10 & - & $93.60$ & $15.50$ \\
     \textbf{ARNe} & \textbf{10} & \textbf{99.43} & \textbf{98.93} & \textbf{17.76} \\
      
     \bottomrule
    \end{tabular}
\end{table}

\subsection{Self-attention for visual reasoning}

\begin{figure}[t!]
    \centering
    \includegraphics[trim={0 0 0 1.5cm},clip,width=\textwidth]{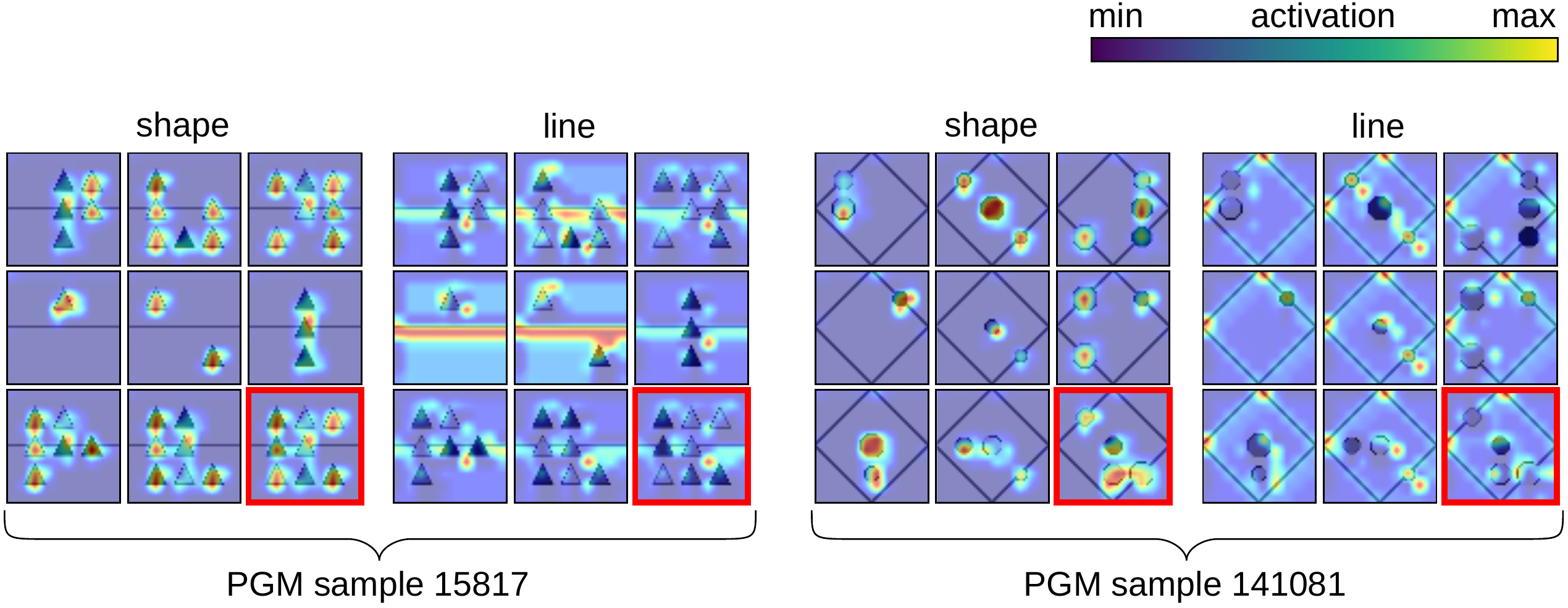}
    \caption{Sum of selected activations of the last layer of the panel encoder CNN, that encoded predominantly for shapes (maps 14, 24, 26, 29, 31) and for lines (maps 1, 2, 3, 4, 8, 12). Matrices show complete sets of context panels and the correct choice panel highlighted in red.}
    \label{fig:attention}
\end{figure}

In order to gain insights into how the self-attention mechanism of \arne{} works internally we visualize the activations of the panel encoder for different encodings of the meta-targets in Figure~\ref{fig:attention}.
It is important to note that these features captures all relevant information for reasoning about relational rules between objects. Therefore, recognizing objects of specific types is an important sub-task for deciding which objects are relevant for a particular relational rule. Figure~\ref{fig:attention} shows two examples for the \emph{object type} meta-targets \emph{shape} and \emph{line}. For each object type, output channels of the panel encoder CNN were selected that strongly responded. The eight choices and the corresponding correct context panel in the bottom right, are shown.

We found that, after training, many panel encoder CNN output channels significantly encoded activation maps for a single meta-target type. The activations in Figure~\ref{fig:attention} most saliently select regions within the input panels that are part of line or shapes objects. 




\subsection{Ablation}

\begin{table}[t!]
\centering
\caption{Comparison of different model configurations. The highlighted lines refer to the best performing model of the given dataset ratio. Values in the training, validation and test columns characterize the model's accuracy. (lr = learning rate). \label{tab:ablation_wren_transformer}}
    \vspace{3mm}
    \small
\begin{tabular}{ccccccccc}
 \toprule
 \textbf{Dropout [\%]} & \textbf{lr\small{$\times10^{-4}$}} & \textbf{Delimiter} & $\beta$ & \textbf{Ratio [\%]} & \textbf{Train [\%]} &  \textbf{Val. [\%]} & \textbf{Test [\%]}\\
 \midrule
10 & 0.5 & \xmark & 10 & 35 & 81.83 & 78.67 & 78.32 \\
10 & 0.5 & \checkmark & 10 & 35 & 82.21 & 79.39 & 78.69 \\
17 & 0.5 & \xmark & 10 & 35 & 93.74 & 85.10 & 84.35 \\
\textbf{17} & \textbf{0.5} & \checkmark & \textbf{10} & \textbf{35} & \textbf{93.83} & \textbf{87.65} & \textbf{87.64} \\
\midrule
10 & 0.5 & \xmark & 0 & 100 & 12.50 & 12.65 & 12.55 \\
17 & 0.5 & \xmark & 10 & 100 & 86.27 & 87.72 & 87.11 \\
10 & 1 & \xmark & 10 & 100 & 88.86 & 88.66 & 87.95 \\
17 & 0.5 & \checkmark & 10 & 100 & 88.23 & 88.42 & 88.04 \\
\textbf{10} & \textbf{0.5} & \xmark & \textbf{10} & \textbf{100} & \textbf{89.06} & \textbf{88.77} & \textbf{88.18} \\
\midrule 
\multicolumn{5}{l}{Feed-forward net instead of Encoder} & 32.02 & 35.87 & 35.06 \\
\multicolumn{5}{l}{Feed-forward net instead of Self-attention} & 38.79 & 45.12 & 44.56 \\
\bottomrule
\end{tabular}

\end{table}


We experimented with various configurations of \arne{}. This involves dropout to reduce over-fitting, a different learning rate, the inclusion of the delimiter token when assembling sequences and the weighting of meta-targets $\beta$.
Table~\ref{tab:ablation_wren_transformer} depicts the results of the subsequent and more detailed ablation studies of ARNe. Further values are as listed in the Appendix~\ref{sec:apx-arne}.
For small dataset dropout gives a substantial performance boost, while this is not the case when the full dataset is used. This is likely due to the additional samples preventing the network from over-fitting.
Similarly, the addition of a deliminiter tends to improve performance for small datasets, thus it might act as a regularizer.
Furthermore, we find that the method is robust to a changed initial learning rate.

An intriguing insight is that \arne{} does not converge when no meta-targets are provided ($\beta=0$). This indicates the relevance of the auxiliary signals which connect the panel choice with the underlying implicit relational rules during training (at test no meta-targets were presented). 
Beyond the findings of Table~\ref{tab:ablation_wren_transformer}, we investigated improvements of components of WReN as well as the addition of FiLM \cite{perez2018film} layers but did not notice a performance improvement.
However, by reducing the input image size from $160\times160$ to $80\times80$ pixels, accuracy decreased by about $15$~ppt. This suggests that relevant small structures exist and are used by the model to solve the PGM.

\subsection{RAVEN Dataset}

For comparison, we also evaluated our model on the recently published RAVEN dataset \cite{zhang2019raven}. We found it achieve a performance of 92.23\% for 50,000 samples for each figure configuration. However, using Raven-10000, i.e. using only 10,000 samples, the performance dropped to 19.67\%.



\section{Conclusion}
\label{sec:conclusion}

In this work we introduced \arne{}, a new deep learning model that combines features from the Wild Relation Network (WReN) and the Transformer network to discover patterns in progressive matrix panels using aspects of fluid intelligence. More precisely, \arne{} builds upon the WReN and extends it with the attention mechanism of the Transformer, which originates from language modelling, making it to our knowledge the first deep learning approach that uses self-attention specifically for abstract visual reasoning. The learning is driven by an auxiliary loss that gives hints about the underlying patterns of the progressive matrix panels. In an extensive experimental comparison we find that \arne{} outperforms state-of-the-art abstract reasoning methods on the PGM dataset by a large margin. Moreover, the analysis shows that our model is substantially more sample-efficient than competing approaches.

Our experiments also including the first application of MAC to the PGM dataset. But the MAC did not seem to be well suited for this task, and also showed significantly worse performance compared to baseline, when it was augmented with a WReN-based panel encoder. Additional experiments are needed to gain a deeper understanding on why MAC appears to fail here. First insights into the inner workings of \arne{} are given in Figure~\ref{fig:attention}. We show that the panel encoder CNN generates meaningful features that aid the learning goal. This suggests that the WReN and Transformer network parts of \arne{} learn to efficiently cooperate for abstract reasoning.

Altogether, these results suggests that the self-attention mechanism can be helpful in domains beyond text processing. Future work involves modifying our method such that it can be used on visual reasoning datasets which require a different structure, e.g. VisualQA datasets which involve parsing an explicit question.

\bibliographystyle{iclr2020_conference}

\newpage
\appendix

{\huge \textsc{Appendix}}


\section{Details to ARNe Implementation}

\label{sec:apx-arne}

The model was trained using a learning rate of $0.5\cdot 10^{-4}$. A heuristic learning rate scheduler was employed which is based on the official MAC implementation \citep{mac_github}. It triggers below a loss of $0.6$ and utilizes a decay parameter of $0.75$. Early stopping \citep{Goodfellow-et-al-2016} was used which stopped training after three subsequent epochs of no improvements of the model's validation loss. For optimization, Adam \citep{kingma2014adam} was used. Parameters of \arne{}'s Transformer encoder are listed in Table~\ref{tab:model_encoder_transformer}.
We used our own PyTorch implementation of the Transformer since at the beginning of this project the reference implementation\footnote{\url{https://github.com/pytorch/pytorch/releases/tag/v1.2.0}} was not available.

\begin{table}[hb]
    \centering
    \caption{Model parameters of the Transformer encoder used in the conducted experiments.}
    \label{tab:model_encoder_transformer}
\begin{center}
\begin{tabular}{ll} 
\toprule
 Dropout Layer & $\text{p}_\text{drop}$ $\left[ \% \right]$\\ 
 \midrule
  $\text{drop}_\text{attention}$ & 10  \\ 
 $\text{drop}_\text{position}$ & 10 \\
 \bottomrule
 \toprule
 Parameter & value \\
 \midrule
 $d_\text{model}$ & 512 \\
 $d_\text{k}$ & 64 \\
 $d_\text{q}$ & 64 \\
 $d_\text{v}$ & 64 \\
 h & 10 \\
 $N_\text{layers}$ & 6 \\
 $d_\text{hidden}$ & 2056 \\
 \bottomrule
 \toprule
 Linear Layer & value \\
 \midrule
 $\text{linear}_\text{v}$ & $\left[d_\text{model}, h \cdot d_\text{v} \right]$ \\
 $\text{linear}_\text{k}$ & $\left[d_\text{model}, h \cdot d_\text{k} \right]$ \\
 $\text{linear}_\text{q}$ & $\left[d_\text{model}, h \cdot d_\text{q} \right]$ \\
 fc & $\left[h \cdot d_\text{k}, d_\text{model}  \right]$ \\
 \bottomrule
 \toprule
 Convolutional Layers & $\left[d_\text{input}, d_\text{output}, \text{kernel}, \text{stride}, \text{padding} \right]$\\
 \midrule
 $\text{FC}_\text{position}$ & $\left[d_\text{model}, d_\text{hidden}, 1, 1, 0 \right]$\\
 & ReLU \\
 & $\left[d_\text{hidden}, d_\text{model}, 1, 1, 0 \right]$\\
 \bottomrule
\end{tabular}
\end{center}
\end{table}

\section{Details to WReN implementation}

\label{sec:reimplementation_wren}

\begin{figure}[t!]
    \centering
     \includegraphics[width=0.9\textwidth]{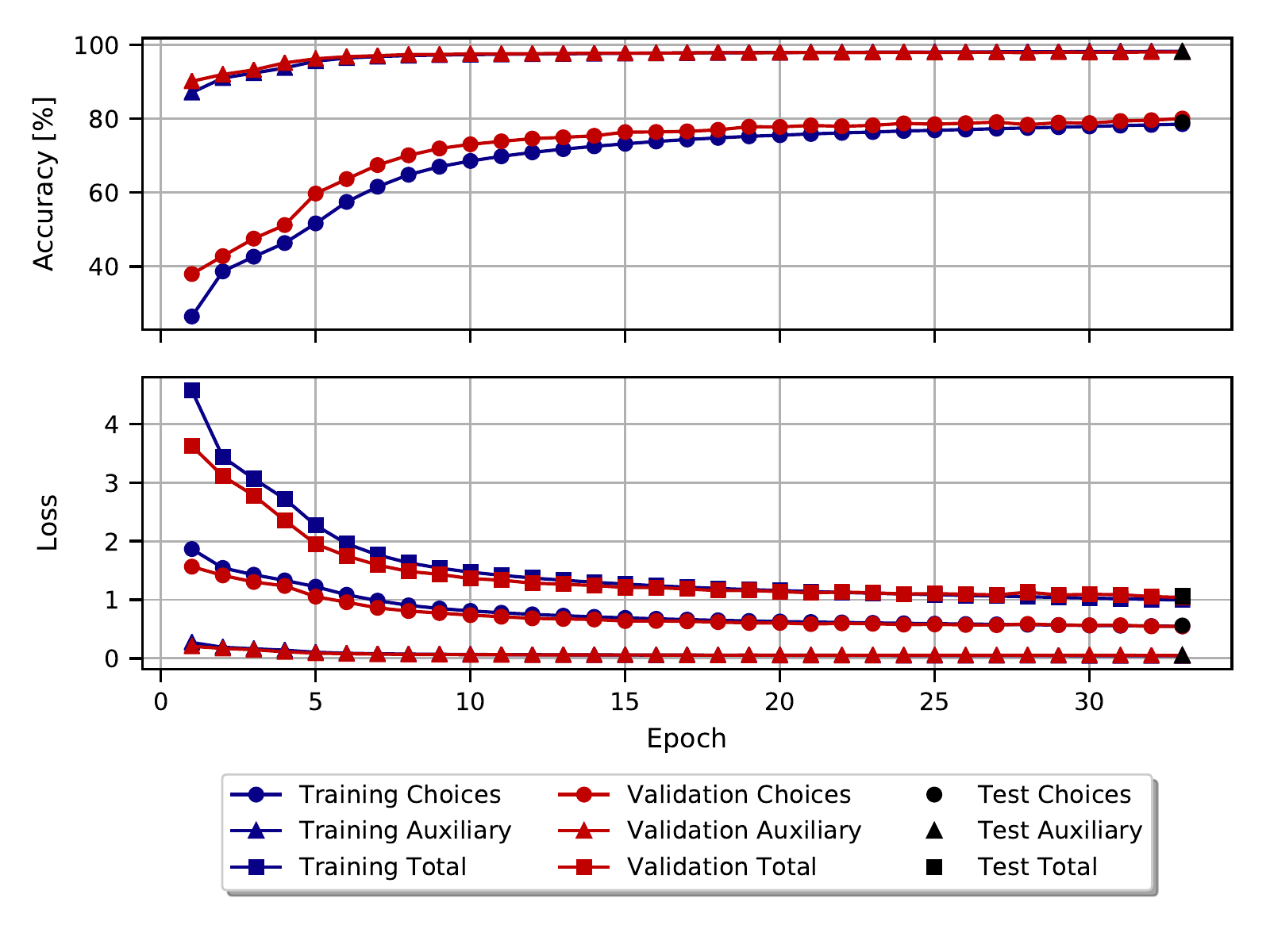}
    \caption{\textit{Accuracy and Loss of the self implemented WReN \citep{barrett2018measuring} model for $\beta=10$ on the neural PGM dataset. The development of both metrics include the progression of the panel choices, additional auxiliary data which describes the setup of the PG-Matrices. In addition, the loss shows the overall sum.}}
    \label{fig:wren_reimplementation}
\end{figure}

The WReN model was re-implemented in PyTorch as described in \citep{barrett2018measuring} and tested on the neutral PGM dataset \citep{pgm_github_deepmind}. The model was trained with $\beta=10$ only. A deviating batch size of $64$ instead of $32$ was used. This yields accuracy rates of $78.49\%$, $80.02 \%$ and $79.00\%$ for training, validation and testing respectively. This implies an improvement of $2.82$~ppt for validation and $2.1$~ppt for testing compared to the WReN baseline model  \citep{barrett2018measuring}.
Figure~\ref{fig:wren_reimplementation} displays the progression of accuracy and loss during training, validation and testing.

\section{Details to MAC implementation}

\label{sec:reimplementation_mac}

\begin{figure}[t!]
    \centering
    \includegraphics[width=\textwidth]{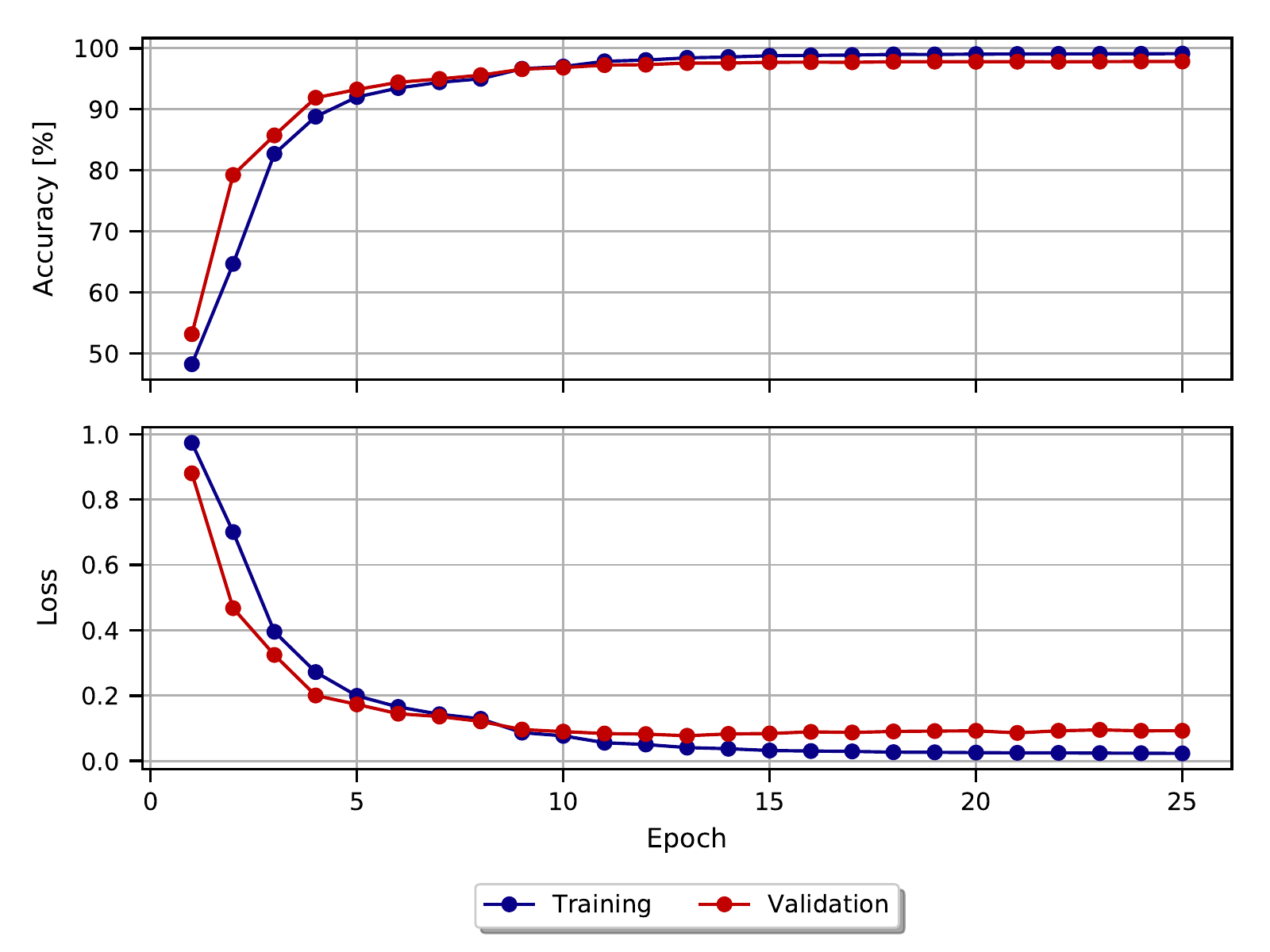}
    \caption{Learning curves during training and validation of our implementation of MAC on the CLEVR dataset. \textbf{Accuracy}: Final learning rate values after $25$ epochs are $97.83\%$ and $99.11\%$ for the validation and training set respectively. \textbf{Loss}: Calculated losses of the implemented neural network with respect to epochs. Final loss values after $25$ epochs are $92.30 \cdot 10^{-3}$ and $22.90 \cdot 10^{-3}$ for the validation and training set respectively.}
    \label{fig:mac_clever_accuracy_loss}
\end{figure}

The MAC network was implemented as reported in \citep{hudson2018compositional} using PyTorch. The implementation was verified on the CLEVR dataset. Loss and accuracies of our implementation of MAC on  the CLEVER dataset can be seen in Figure~\ref{fig:mac_clever_accuracy_loss}. Best value for validation regarding the accuracy is $97.83\%$. The training, validation demanded a computation effort of about half a week on a GeForce GTX TITAN X graphics card.

\section{Details to WReN-MAC Implementation}
\label{sec:wren-mac}

\begin{figure}[t!]
    \centering
    \includegraphics[width=0.8\textwidth]{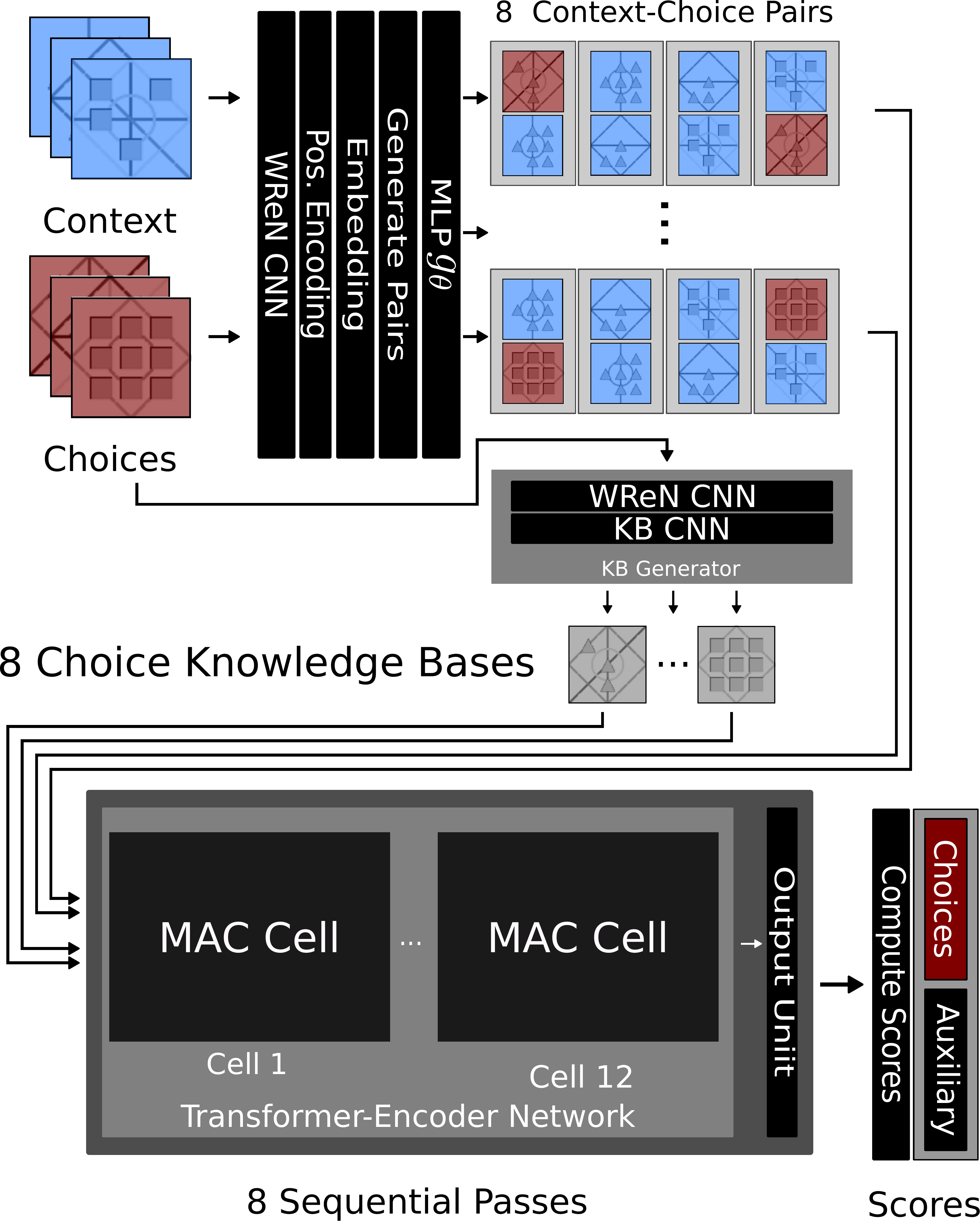}
    \caption{WReN-MAC model. First, eight sequences of embeddings and $g_\theta$ activations are generated analogously to WReN. To interface a MAC-cell properly, the knowledge base is required. A knowledge base is generated by the WReN CNN plus additional layers to match MAC's required dimensions. Every sequence passes 12 recurrent MAC-cells and the output unit sequentially. The computation of scores equals the method incorporated in the WReN model.}
    \label{fig:wren_mac_model}
\end{figure}

The WReN-MAC model uses the same encoding mechanism as the WReN baseline model and includes 12 MAC-cells. In order to interface a MAC-cell, additional convolutional layers were appended to the adapted WReN convolutional network. The model is analogously sequentially aligned like WReN or WReN-Transformer whereas during each pass one knowledge base which encodes one distinct choice panel is used. The question vector $\vec{q}$ is computed in analogy to WReN where $g_\theta$ and the subsequently applied sums were used. The computation of the model's loss also respected auxiliary structure set data of the corresponding PGMs. 

\remark{duplicate at the end of paragraph: radient clipping; scheduler}
We were not able to train WReN-MAC successfully on the PGM dataset. Early stopping finished learning after $33$ epochs. The final accuracies and losses are $49.26 \%$ and $3.04$, $46.97\%$ and $3.20$, $46.89\%$ and $3.23$ for training, validation and testing respectively. Both accuracy and loss showed marginal improvements throughout all epochs.
Due the recurrent nature of MAC-cells, we used gradient clipping \citep{Goodfellow-et-al-2016}. The learning rate scheduler was set but didn't diminished the the learning rate which was initially set to $1 \cdot 10^{-4}$. Additional hyperparameter tuning did not show significant performance increase.

\end{document}